\documentclass[10pt,twocolumn,letterpaper]{article}

\usepackage{cvpr}
\usepackage{times}
\usepackage{epsfig}
\usepackage{graphicx}
\usepackage{amsmath}
\usepackage{amssymb}
\usepackage{gensymb}
\usepackage{xcolor}
\usepackage{caption}
\usepackage{subcaption}
\usepackage{textcomp}

\usepackage[pagebackref=true,breaklinks=true,letterpaper=true,colorlinks,bookmarks=false]{hyperref}

\cvprfinalcopy 

\def\cvprPaperID{18} 

\ifcvprfinal\fi

\makeatletter
\def\blfootnote{\xdef\@thefnmark{}\@footnotetext}
\makeatother

\begin{document}

\title{Probabilistic Oriented Object Detection in Automotive Radar}

\author{Xu Dong\textsuperscript{*} \qquad Pengluo Wang\textsuperscript{*}  \qquad Pengyue Zhang \qquad Langechuan Liu \textsuperscript{$\dagger$} \\
XSense.ai\\
}

\maketitle
\thispagestyle{empty}

\blfootnote{\textsuperscript{*} indicates equal contributions.}
\blfootnote{\textsuperscript{$\dagger$} {indicates corresponding author \tt patrickl@xsense.ai}}

\begin{abstract}
Autonomous radar has been an integral part of advanced driver assistance systems due to its robustness to adverse weather and various lighting conditions. Conventional automotive radars use digital signal processing (DSP) algorithms to process raw data into sparse radar pins which do not provide information regarding the size and orientation of the objects. In this paper we propose a deep-learning based algorithm for radar object detection. The algorithm takes in radar data in its raw tensor representation and places probabilistic oriented bounding boxes (oriented bounding boxes with uncertainty estimate) around the detected objects in bird's-eye-view space. We created a new multimodal dataset with 102,544 frames of raw radar and synchronized LiDAR data. To reduce human annotation effort we developed a scalable pipeline to automatically annotate ground truth using LiDAR as reference. Based on this dataset we developed a vehicle detection pipeline using raw radar data as the only input. Our best performing radar detection model achieves 77.28\% AP under oriented IoU of 0.3. To the best of our knowledge this is the first attempt to investigate object detection with raw radar data for conventional corner automotive radars.

\end{abstract}

\section{Introduction}
Object detection holds the key for achieving autonomous driving. While camera and LiDAR have been the two major sensory modalities in autonomous driving field, they both have their own drawbacks. For example, 3D object detection from camera alone proves to be very challenging so far despite recent progress \cite{wang2019pseudo}, and LiDAR is inherently not reliable in adversarial driving conditions \cite{bijelic2018benchmark, kutila2018automotive} and still too expensive for mass production. On the other hand, radar, as a widely-adopted sensor in traditional Advanced Driver Assistance Systems (ADAS), is very robust and reliable under different weather conditions. Using radar for object detection can be of great help for increasing both redundancy and robustness of perception in autonomous driving.

\begin{figure}[t]
\begin{center}
  \includegraphics[width=1\linewidth]{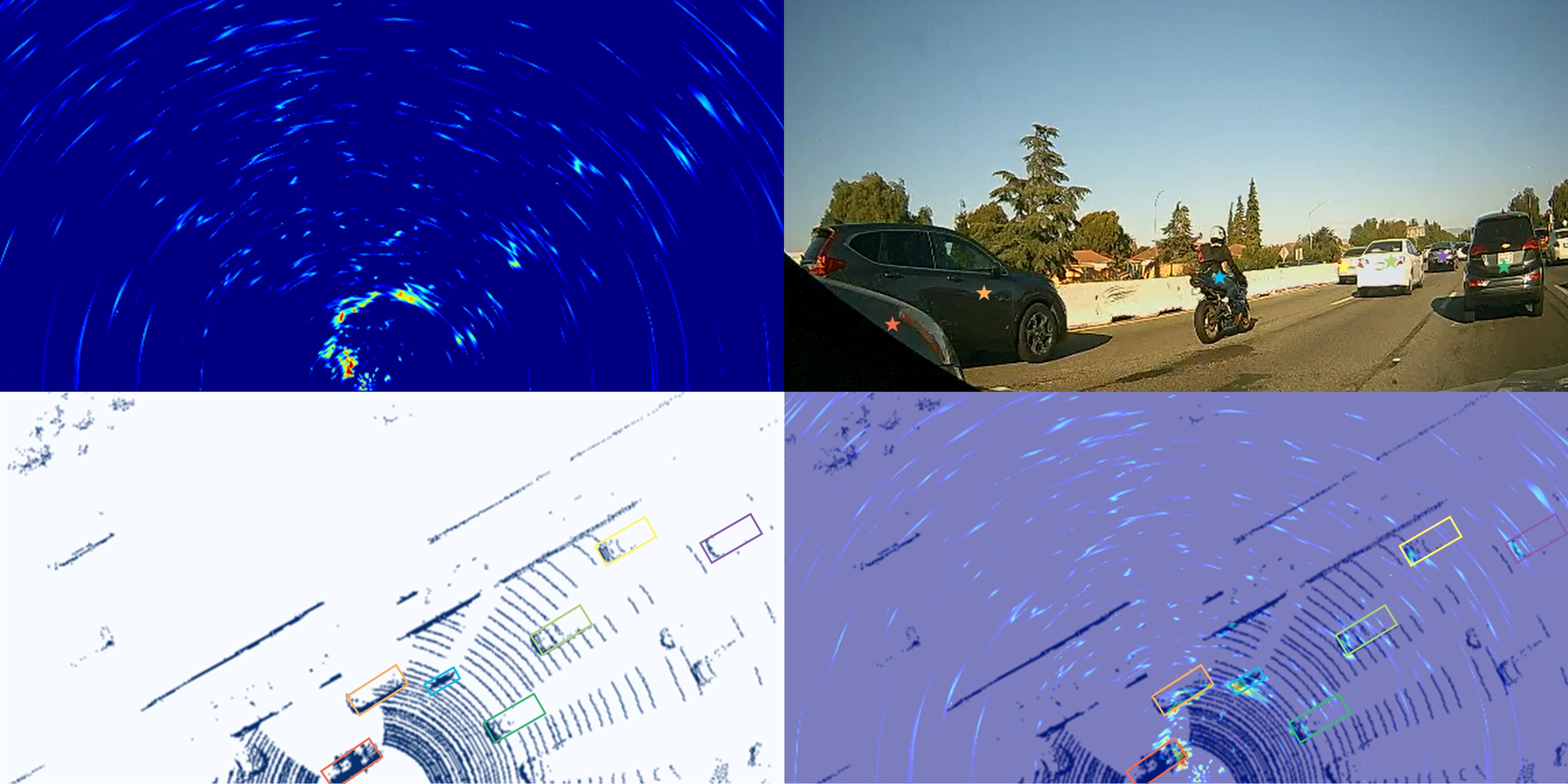}
\end{center}
  \caption{An sample scene from our dataset, with radar, camera, LiDAR, and radar-LiDAR overlaid data. The camera image is only used for visualization and is not consistently collected in our dataset. Radar is mounted at the front left corner at bumper height and LiDAR on the rooftop. The top left image is the radar data in data tensor representation; top right is the corresponding camera image; bottom left is the corresponding LiDAR data in bird's-eye-view (BEV) image; bottom right is the overlaid image from radar and LiDAR data, for better understanding of the semantics. Objects in the scene are marked with colored stars in the camera image, and colored boxes in the LiDAR image and overlaid image.}
\label{fig:sensor_demo}
\end{figure}

Data from radar can take on different representations. For conventional automotive radars, raw data are heavily processed by digital signal processing (DSP) algorithms and are reduced to sparse radar pins (normally only 10 to 50 points per frame). Under this representation, one object is generally denoted by only one radar pin, without any size and orientation information. One emerging revolution in the industry is \textit{imaging radar}, which can produce semi-dense radar point cloud in a similar format to LiDAR point cloud \cite{astyx_dataset}. Compared with sparse radar pins, point cloud representation requires fewer DSP modules and contains more low-level information. However, similar to LiDAR sensors, the \textit{imaging radar} are more costly and not fully ready for deployment in mass production cars.

Another trend in autonomous driving is the use of corner radars for 360$\degree$ surveillance. Compared with front radar, object detection using corner radars has more challenges due to multipath propagation and a shorter detection range. Fig. \ref{fig:sensor_demo} shows what corner radar signal looks like under radar tensor representation together with synchronized LiDAR and camera data.

In this study, we choose to work with the widely available conventional (\textit{i.e.}, non-imaging) radar but with a novel radar tensor representation. Under this representation, minimum radar DSP is applied to preserve maximum information in the raw data. We leverage the recent advances in object detection and Bayesian deep learning to extract information from the intrinsically noisy \cite{shnidman2003expanded} radar data.  Compared with radar pins, radar tensor representation makes it possible to predict additional information such as the class, size, and orientation. This information is fundamental for autonomous driving system and can be consumed directly by downstream modules such as sensor fusion.

A big obstacle for performing object detection on radar data is the lack of public dataset. To the best of our knowledge, for all autonomous driving datasets, only nuScenes \cite{nuscenes} provides radar data, but still under traditional radar pin representation. Therefore, we built a new dataset collected under realistic autonomous driving scenarios. This dataset contains raw radar tensor and LiDAR point cloud as well, which can potentially serve for multimodal detection purposes. In this study we only focus on radar-based object detection using convolutional neural networks (CNN), and our goal is to predict object class, location, size, and orientation with radar input only.

Our main contributions in this work are summarized as:
\begin{itemize}
    \item Proposition of a radar tensor representation which is more suited for CNN based object detection algorithm compared to other traditional radar representations. To the best of our knowledge this is the first attempt to investigate object detection with conventional (non-imaging) corner radar under this representation.
    \item Development of a scalable pipeline for data curation. Ground truth is automatically generated using LiDAR data to minimize human annotation effort. We also designed some techniques such as LiDAR re-packing to improve cross sensor calibration and label accuracy.
    \item Adaptation of CNN-based object detection algorithm for radar data with uncertainty estimation for bounding box localization. The feasibility of radar object detection is demonstrated and an AP of 77.28\% has been achieved under oriented IoU threshold of 0.3.
\end{itemize}


\section{Related Work}

\subsection{Object detection} 
CNN based object detector can be generally grouped into two categories, multi-stage detectors \cite{fast-rcnn, faster-rcnn, cascade-rcnn} and single stage detectors \cite{retinanet, ssd, yolov3}. Multi-stage detectors first generate proposals and then gradually refine detection results to increase detection performance. Single-stage detectors directly output final prediction in a single stage and hence can achieve faster detection speed compared with multi-stage detectors without sacrificing performance too much. The good balance between speed and performance makes single-stage detectors suitable in real-time applications such as autonomous driving. On the other hand, Single-stage detectors can be further categorized into anchor-based \cite{DFP, STDN, R-FCN} and anchor-free algorithms \cite{extremenet, cornernet, centernet-2, foveabox, reppoints}. Anchor-based algorithms use anchor boxes to facilitate the regression of object localization information. The anchor-free algorithms formulate object detection as a keypoint detection problem, so that it does away with the ad-hoc heuristics and handcrafted design of anchors and still demonstrated state-of-the-art performance in the latest researches \cite{centernet, reppoints}. 

\subsection{Oriented Object Detection} 
Oriented object detection has been extensively investigated recently in text region detection \cite{jiang2017r2cnn, ma2018arbitrary}, remote sensing \cite{xia2017dota, liu2016ship} and autonomous driving \cite{mousavian20163d}. Commonly used techniques include deformable convolutional neural networks \cite{dcnn, dcnn-v2} and spatial transformers \cite{spatial-transformer}. Different from the upright anchor boxes in general object detection, oriented anchors are placed to assist the regression of oriented bounding boxes \cite{xia2017dota,liu2016ship,Ma_2018}. In addition, the multi-bin losses are usually used by combining classification and regression for accurate orientation estimation \cite{massa2016crafting,mousavian20163d}.  


\subsection{Probabilistic Object Detection}
Uncertainty estimation in deep learning is first explored by the seminal work \cite{uncertainties}, and then widely adopted in many object detection algorithms \cite{lasernet, gaussianYOLO, monoloco,he2019bounding, Neven_2019_CVPR}. There exist two different uncertainties, namely epistemic uncertainty and aleatoric uncertainty. While epistemic uncertainty can be explained away with more data, aleatoric uncertainty is inherent in sensor data. In this paper, we will introduce the probabilistic model to our detection algorithm to cope with the inherent uncertainty in radar data \cite{shnidman2003expanded}.

\subsection{Radar-based Object Detection} 
Range, azimuth, elevation and velocity are the common attributes to be estimated during radar-based detection, and traditionally this was solved purely by DSP algorithms. For example, Multiple Signal Classification (MUSIC) \cite{music} and Estimation of Signal Parameters via Rotational Invariance Technique (ESPRIT) \cite{esprit} are the common methods for estimating Angle of Arrivals of an object, which is associated with azimuth and elevation information. And in addition, some other adaptive algorithms such as cell-averaging CFAR \cite{cfar, fast-cfar} will be used to detect objects of interest embedded in noise and cluster. However, radar ``object detection''
by these DSP algorithms simply reveals the location (and optionally, velocity) of objects, and should be differed from object detection in computer vision.

In recent years, deep learning based methods start to be investigated on radar signal for object detection with semantic meaning. In \cite{furukawa2018deep}, an end-to-end CNN has been proposed for automatic target recognition using SAR image. Deep Radar Detector \cite{brodeski2019deep} directly uses 4D FMCW radar data containing information of range, Doppler, azimuth, and elevation for object detection. In \cite{Patel2019DeepLO}, CNN is applied to radar spectrum regions-of-interest (ROI), whose location is detected by traditional DSP algorithms. More deep learning based methods remain to be investigated in radar object detection area.

\section{Radar Dataset Buildup}
In this section, we will introduce the data collection and curation processes for building our radar object detection dataset. As it is challenging to obtain object detection ground truth from radar signal alone, we also collected LiDAR data for ground truth generation. We mounted the radar at the front left corner of the vehicle at bumper height and the LiDAR (Velodyne HDL-64E) on top of the roof. Details about the dataset buildup are explained below. 

\subsection{FMCW Radar Signal and Radar DSP}\label{ch:fmcw_desc}
Frequency Modulated Continuous Wave (FMCW) radars are very popular for autonomous driving with the ability to measure \textit{range} (radial distance), \textit{velocity} (Doppler), and \textit{azimuth} information. FMCW radars continuously transmit chirp signal (a sinusoidal signal whose instantaneous frequency varies linearly and periodically) and receive echo signal reflected by objects. The transmitted chirp signal is highly configurable and determines radar signal specifications. Table \ref{tab:fmcw_index} shows the key configurations for our FMCW radar. Due to limited data transfer bandwidth, there exists a trade-off between the ability to measure velocity information and the accuracy of range and azimuth measurements. We decided not to collect velocity measurements in our radar setting in order to achieve best range and azimuth performance.

\begin{figure}
     \centering
     \begin{subfigure}[b]{0.092\textwidth}
         \centering
         \includegraphics[width=\linewidth]{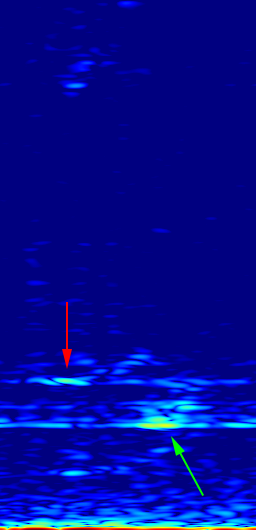}
         \caption{FFT\\(\textit{data\_fft})}
         \label{fig:data_fft}
     \end{subfigure}\hspace{\fill}
     \begin{subfigure}[b]{0.38\textwidth}
         \centering
         \includegraphics[width=\linewidth]{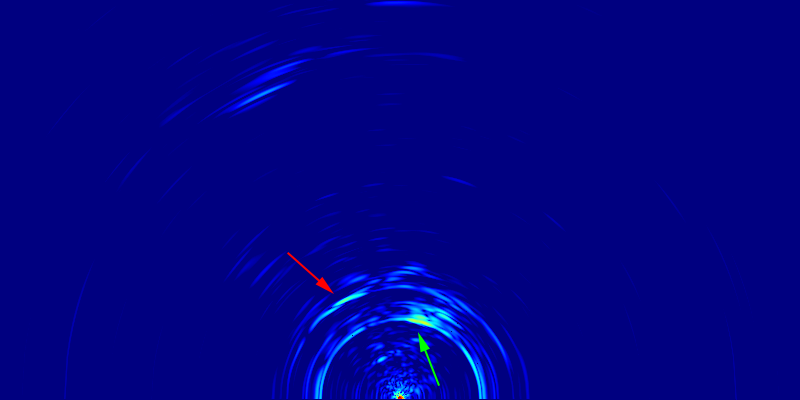}
         \caption{FFT under Cartesian\\(\textit{img\_fft})}
         \label{fig:img_fft}
     \end{subfigure}
     \\
     \begin{subfigure}[b]{0.092\textwidth}
         \centering
         \includegraphics[width=\linewidth]{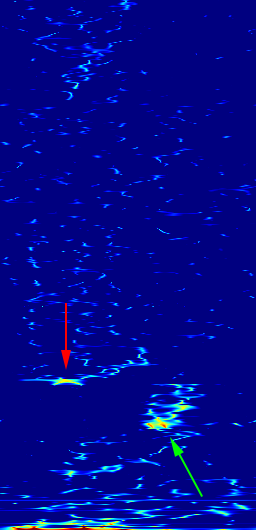}
         \caption{MUSIC\\(\textit{data\_music})}
         \label{fig:data_music}
     \end{subfigure}\hspace{\fill}
     \begin{subfigure}[b]{0.38\textwidth}
         \centering
         \includegraphics[width=\linewidth]{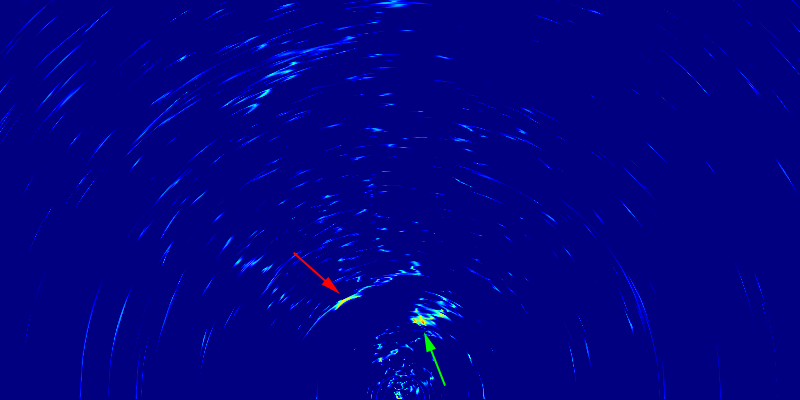}
         \caption{MUSIC under Cartesian\\(\textit{img\_music})}
         \label{fig:img_music}
     \end{subfigure}
        \caption{Different formats of radar data. (a) and (c) are shown in polar coordinate system, with horizontal axis representing \textit{azimuth} from $-90\degree$ to $90\degree$ and vertical axis representing \textit{range} from 0 to 40 m. After coordinate transformation into Cartesian coordinate system, (b) and (d) are BEV images which span 40 meters in forward, leftward, and rightward directions. By comparing results from the two rows, we can easily find out that side lobes are greatly suppressed after using MUSIC algorithm. Green arrows in the plots point out a passing vehicle in front of our corner radar. Red arrows point out the barrier of highway.}
        \label{fig:radar_dsp}
\end{figure}

\begin{table}[t]
\caption{Key configurations for our FMCW radar}
\label{tab:fmcw_index}
\begin{center}
\begin{tabular}{c|c}
    \hline\hline
    \textbf{Attribute}     &  \textbf{Value}\\[1pt]
    \hline
    Maximum range     & 153.60 m \\[1pt]
    Range resolution & 0.15 m\\[1pt]
    Maximum azimuth & $\pm 90 \degree$\\[1pt]
    Azimuth resolution & $3.7 \degree$\\[1pt]
    Frame rate &   50 Hz\\[1pt]
    \hline \hline
\end{tabular}
\end{center}
\end{table}


Raw data collected from radar are 2D array with size $N_s$ and $N_{ch}$, where $N_s$ represents the number of points in analog to digital (A/D) sampling and $N_{ch}$ represents the total number of receiver antennas. Two-dimensional Fast Fourier Transformation (FFT) was then used to obtain range-azimuth data \cite{acc_radar_sim}. Since range-azimuth data is under polar coordinate system, coordinate transformation has been implemented to generate pseudo image under Cartesian coordinates. Results before and after coordinate transformation are shown in Figs. \ref{fig:data_fft} and \ref{fig:img_fft}. However due to a relatively small number of receiver antennas (in our case, $32$), the side lobes caused by FFT on azimuth dimension are manifested as artifact (highlighted horizontal lines on range-azimuth map and rings under Cartesian coordinates), which obscure actual object locations as shown in the figures. In order to solve this problem, MUSIC \cite{music} has been used as a super-resolution algorithm to suppress side lobes, with results shown in Fig. \ref{fig:data_music} and \ref{fig:img_music}. 

By using different methods, four different formats of radar data, namely, data-fft (with FFT only), img-fft (with FFT and coordinate transformation), data-music (with MUSIC only), and img-music (with MUSIC and coordinate transformation) are visualized in Fig. \ref{fig:radar_dsp}.

\subsection{Calibration and Synchronization}
As can be seen from Fig. \ref{fig:radar_dsp}, labeling with only radar input is extremely difficult. Therefore, LiDAR data have to be used as reference to obtain ground truth information. Successful transfer of annotation from LiDAR to radar requires good synchronization and cross-sensor calibration. We refer to the aligned radar and LiDAR images as \textit{image pairs} hereafter.

\textbf{Spatial Calibration:} Radar and LiDAR data are under different coordinate systems due to different mounting positions. Calibration between radar and LiDAR are implemented by using trihedral radar reflectors as landmarks and calculating coordinate transformation parameters from the correspondences of the landmarks in image pairs.

\begin{figure}[t]
\begin{center}
  \includegraphics[width=.8\linewidth]{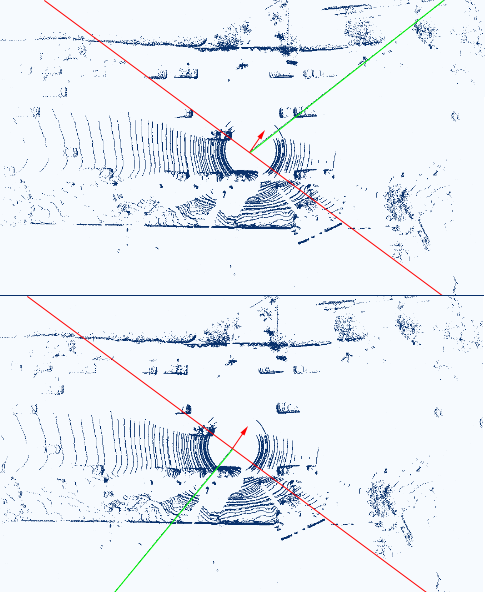}
\end{center}
  \caption{An example for rolling shutter effect and its solution. Here LiDAR point cloud data are rendered in BEV space. Radar field of view is the top right half in each image separated by the red line and indicated by the red arrow. The green straight line originated from center of the image is the initial scan angle of current LiDAR sweep. As can be seen in the upper figure, the vehicle located at the green line is abnormally elongated. After re-packing, the artifact is removed.}
\label{fig:lidar_shutter}
\end{figure}

\textbf{LiDAR Re-packing:} Pronounced rolling shutter effect is found in LiDAR data around the initial scan angle at each LiDAR sweep. Fig. \ref{fig:lidar_shutter} displays an example of this effect, where we can see a car is elongated at the initial scan angle, inducing ambiguity of its position. This effect is common when vehicles come from the opposite direction and thus have higher relative velocity with the ego car. Due to time delay caused by rotating $360 \degree$ of LiDAR laser, passing vehicles will have non-negligible position offset. When initial scan angle of a LiDAR sweep is within the radar field of view, such ambiguity will complicate radar-LiDAR synchronization process. To tackle this problem, we re-packed the LiDAR sweeps to move initial scan angle out of the radar field of view. The corresponding re-packed LiDAR sweeps do not have this effect as shown in the figure. 

\textbf{Temporal Synchronization:} After cross-sensor calibration and LiDAR re-packing, radar and LiDAR image can be overlaid perfectly as long as two sensors are synchronized, which is done by two steps. Firstly, after LiDAR re-packing, new frames are generated with shifted initial scan angle, thus timestamp for each new frame is re-calculated by linear interpolation of the timestamps of two previous neighboring frames. Secondly, due to internal clock difference, we manually find the time offset for each data collection to achieve a best match of overlaid radar and LiDAR images. The radar and LiDAR data are collected at 50 Hz and 10 Hz, and after temporal synchronization, they are associated as 10 Hz image pairs.

\subsection{Data Auto Labeling}
Obtaining ground truth for radar data requires tremendous manual annotation effort by looking at image pairs. To reduce the human annotation effort, we developed an auto-labeling procedure, with two steps explained as follows. 

The first step is to perform object detection on LiDAR data with the goal to achieve as high detection recall as possible. We combined two state-of-the-art LiDAR object detection algorithms, namely ComplexYOLO \cite{complexyolo} and PointRCNN \cite{pointrcnn}, and also used test time augmentation and model ensemble to further boost detection performance. For test time data augmentation, we combined 90-degree rotation around z-axis (up-down) with horizontal mirroring around y-axis (left-right with respect to ego car heading) in LiDAR coordinates, yielding a total of eight different LiDAR frames from one frame. In combination of the two detection models, 16 sets of detection results would be generated from one LiDAR frame. Soft NMS with threshold 0.9 is used to fuse those 16 sets into final prediction.

The second step is to transform LiDAR object detection results into radar coordinate system, and filter out false positive detection results in order to increase precision. To reduce false positive detections, we calculated the strength of radar signal response in each detection region, and discard the detection with low responses. Concretely, we characterized response strength by calculating the Area under the Curve (AUC) of the normalized cumulative density function formed by values of all pixels in the region of each detection. The region was enlarged by 20\% of the original detected bounding box due to the fact that radar response of a an object mostly happens on the boundary. With such auto-labeling procedure, ground truth for radar object detection can be generated for massive radar dataset in a scalable way.


\subsection{Dataset Overview}
\begin{figure}[t]
\begin{center}
  \includegraphics[width=\linewidth]{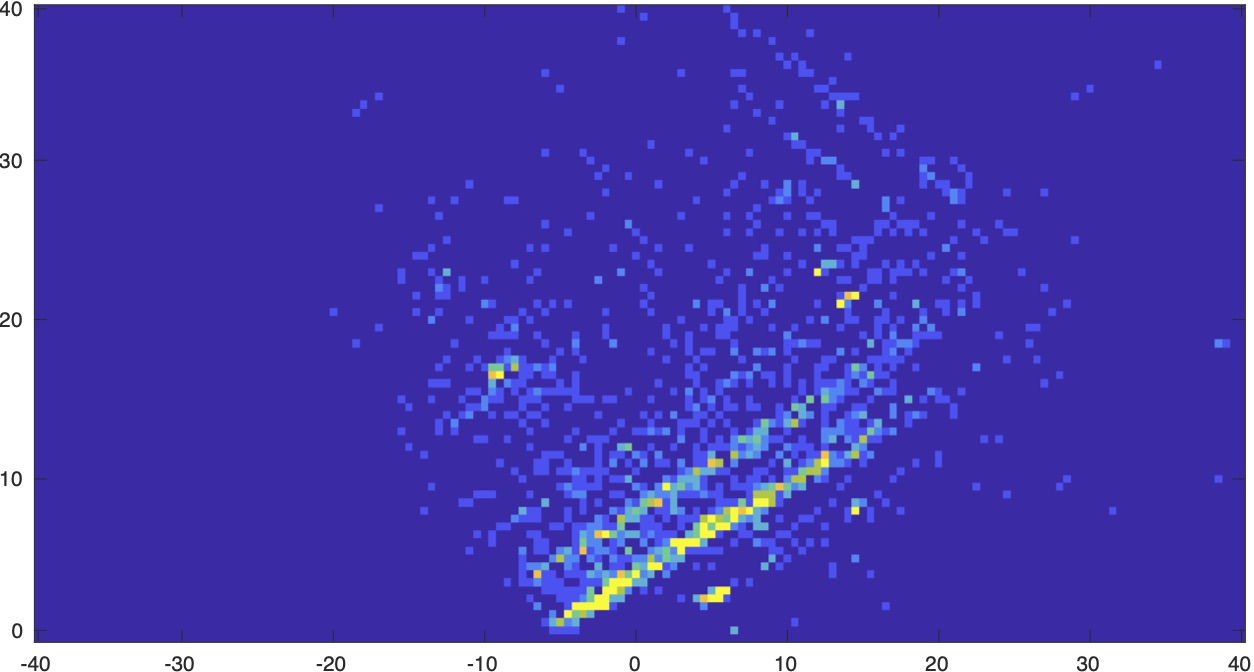}
\end{center}
  \caption{Spatial distribution of ground truth bounding boxes.}
\label{fig:ran_azim_dist}
\end{figure}

We further downsample the synchronized radar and LiDAR data from 10 Hz into 2 Hz to reduce temporal correlation between adjacent frames. The final radar dataset contains 102,544 images in total. Fig. \ref{fig:ran_azim_dist} shows the spatial distribution of ground truth bounding boxes. Besides the auto-labeling process described above, we also manually verified 25\% of all images. Manually verified data were categorized into different scenes and the distribution is displayed in Table \ref{tab:scene_dist}.

\begin{table}[]
\caption{Scene distribution}
\label{tab:scene_dist}
\begin{center}
\begin{tabular}{c|c|c|c|c}
    \hline\hline
    \textbf{Highway}&\textbf{Local}&\textbf{Crossing}&\textbf{Parking-lot}&\textbf{Other}  \\[1pt]
    \hline
    22\%&47\%&22\%&7\%&2\%  \\[1pt]
    \hline \hline
\end{tabular}
\end{center}
\end{table}

\section{Object Detection on Radar Data}
\begin{figure*}[t]
\begin{center}
  \includegraphics[width=0.9\linewidth]{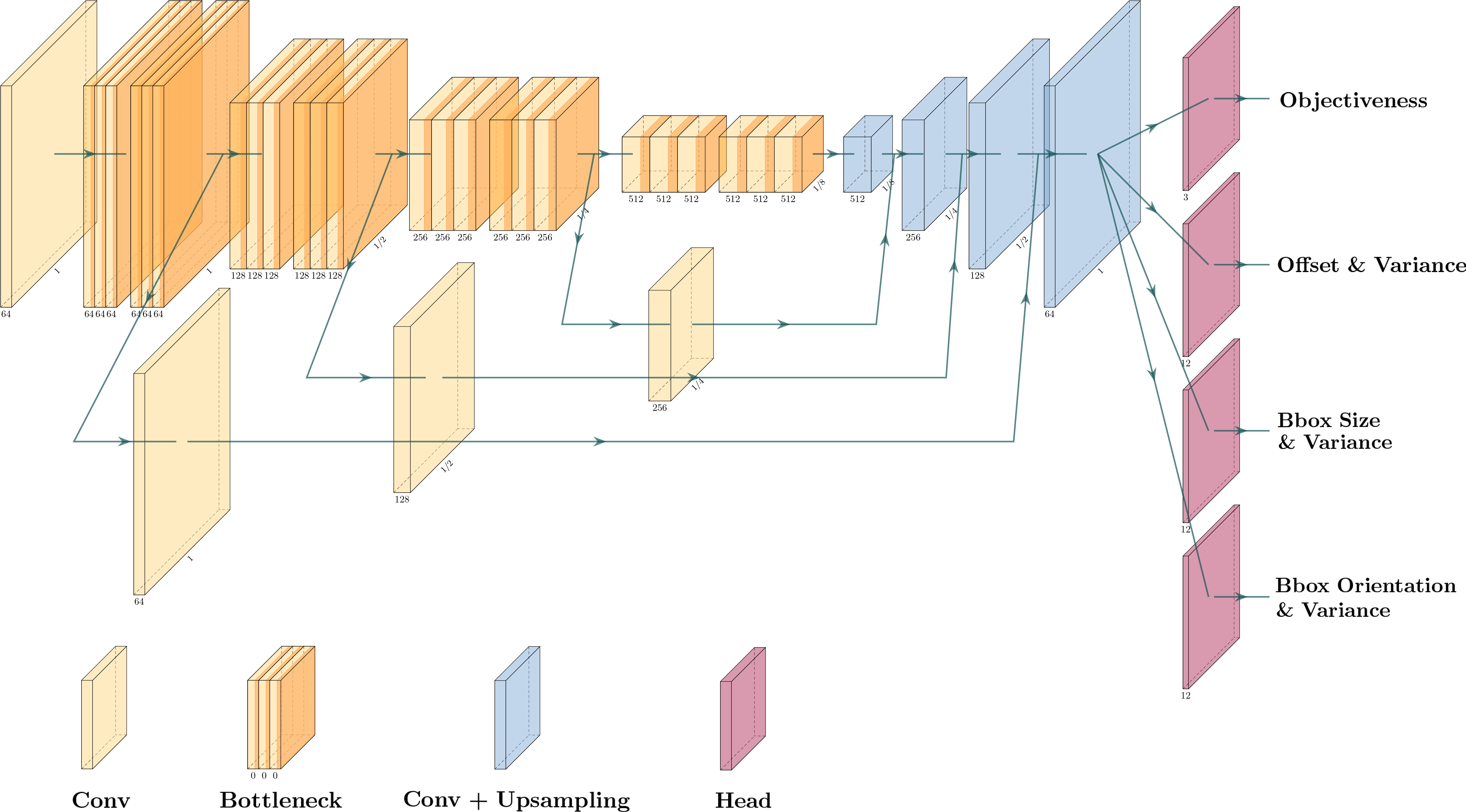}
\end{center}
  \caption{Network structure of our detection algorithm. The encoder of the network is adapted from ResNet-18 and has five stages, where the first stage contains a conv7$\times$7 layer, and each of the following four stages contains two bottleneck blocks. The decoder has four stages with each stage consisting of two conv3$\times$3 layers. The encoder and decoder are laterally connected at each stage with a conv1$\times$1 layer. Downsampling in the encoder is achieved by conv3$\times$3 with stride size as 2$\times$2, and the upsampling in the decoder is achieved by bilinear upsampling operation. The predictions are generated from the last feature map followed by two conv3$\times$3 layers.}
\label{fig:complexyolo}
\end{figure*}

We formulate the task of car detection with radar data as a probabilistic oriented object detection problem. As we do not distinguish different types of cars in our algorithm, the task is hence formulated as class agnostic object detection problem. We used single-stage anchor-based detection algorithm in this study, for the sake of trade-off between detection performance and detection speed. Typical single-stage anchor-based object detector usually consists of a backbone, detection neck, and detection head. In our study, for the backbone, the bottleneck block \cite{resnet} was used as the basic unit, and shortcut connections between encoder and decoder blocks are also used to combine the low-level and high-level features together. For the detection neck, we simply used two 3$\times$3 conv layers to connect the backbone with the detection head. Fig. \ref{fig:complexyolo} illustrates the architecture of the network. For the detection head, details are explained as follows.

\subsection{Detection Head}
\textbf{Anchors:} Three different oriented anchors are placed at each grid cell (\textit{i.e.}, pixel in the spatial dimensions) in the network's last feature map. The configuration of the three anchors are determined by the statistics of all ground truth bounding boxes (referred as \textit{gt bboxes} hereafter) via k-means clustering. Concretely, since the sizes of gt bboxes are clustered at one peak position (due to BEV projection and the unbalanced vehicle classes in our dataset), all anchors are set to be the same as the mean size of all gt boxes ($15 \times 42$ pixels). Similarly, instead of using multi-scale feature maps for predictions as in \cite{fpn}, we only used the last feature map. The preset orientations of the anchors are set to be the three clustered orientations of all gt bboxes (30, 115, and 136 degrees). 

\textbf{Predictions:} Each of the three anchors per grid cell in the network's last feature map predicts an oriented bounding box consisting of an objectiveness score ($s$), position ($x$, $y$), size ($w$, $h$) and orientation ($\theta$). The objectiveness score $s \in [0, 1]$ indicates the existence of a bbox from this anchor. During network training, one gt bbox is only assigned to one anchor which matches the gt bbox best. The bbox localization information ($x$, $y$, $w$, $h$, and $\theta$) will be regressed from the anchor if it is assigned to a gt bbox.

\subsection{Localization Regression}
\textbf{Position and Size}: As for regressing the bbox position and size information ($x$, $y$, $w$, and $h$), the direct outputs from the network are the relative offsets ($x_{o}$, $y_{o}$, $w_{o}$, and $h_{o}$) between the anchor and the corresponding gt box, the same as the standard bbox parameterization from R-CNN \cite{rcnn}. After $x$, $y$, $w$ and $h$ of predicted bboxes are calculated from the relative offsets, they are compared with corresponding gt bboxes and supervised with smoothed $L1$ loss ($SL_1$).

\textbf{Orientation}: The network regressed the relative angular offset between the anchor and the corresponding gt box. We transform this angular offset ($\theta_o$) into the corresponding sine and cosine values ($\cos \theta_{o}$, $\sin \theta_{o}$) as the regression targets. During training the outputs from the network are compared to the regression targets and supervised with smoothed $L1$ loss. During inference, the orientation of the bbox is recovered from the anchor's orientation and the predicted angular offset calculated from $\arctan(\cos \theta_{o}, \sin \theta_{o})$.

\subsection{Probabilistic Detection}
Due to the noisy nature and inherent ambiguity in radar signal \cite{shnidman2003expanded}, it is impractical to require neural network to make accurate predictions. However, unlike the objectiveness score which lend itself to probabilistic interpretation and uncertainty estimation, bbox localization predictions ($x$, $y$, $w$, $h$, $\theta$) are deterministic values with no uncertainty estimation. To address this problem, we make the network learn the aleatoric uncertainty for its localization prediction based on \cite{uncertainties}. Specifically, the network will predict the individual variance for each of the the position predictions $(x_{o}, y_{o})$, the size predictions $(w_{o}, h_{o})$, and the orientation predictions $\left(\cos \theta_{o}, \sin\theta_{o}\right)$. And the loss function for each of them is expressed as:


\begin{equation}
L_{a}= \sum_{i=0}^{N} \left(\frac{1}{\sigma_a}  SL_1(a^{pred} - a^{gt}) + \log \sigma_a\right) \tag{1} \label{eq:1}  
\end{equation}
where $a$ represents the prediction of $x_{o}$, $y_{o}$, $w_{o}$, $h_{o}$, $\cos \theta_{o}$, and $\sin \theta_{o}$; $\sigma_a$ is the variance of its prediction; and $N$ is the total number of bounding boxes in a image.

\subsection{Network Loss}
The total loss $L_{total}$ of the network consists of the objectiveness loss $L_{obj}$ and the localization loss $L_{loc}$. For objectiveness loss, we used focal loss \cite{retinanet} with alpha of 0.25 and gamma of 2. The localization loss consists of a smooth L1 loss attenuated by a learned variance and a regularization term proportional to the logarithm of the variance, as in Equation \ref{eq:1}. And total loss is a weighted sum of the objectiveness loss and the localization loss.
\begin{equation}
L_{tot}= L_{obj} + w_0 \sum_{a}L_{a} \tag{2} \label{eq:2}  
\end{equation}
where $a$ $\in$ $\{x_{o}, y_{o}, w_{o}, h_{o}, \cos \theta_{o}, \sin \theta_{o}\}$\textbf{}, and $\sum_{a}L_{a}$ denotes the localization loss. We set $w_0$ to be 100 in our implementation.

\subsection{Training and Inference}
The network was trained in a single NVIDIA GeForce RTX 2080 Ti, with batch size to be 4 using Adam optimizer for 500,000 iterations. The learning rate was set to be 0.0005 initially, and cosine learning rate scheduler \cite{cosinelr} was used to adjust the learning rate with a period of 100,000 iterations. During inference, prediction confidence score was thresholded by 0.5 and non-maximum suppression was used with IoU threshold as 0.0001, because no two cars should overlap in BEV projection in reality. We evaluated the performance of our algorithm by average precision (AP) with three IoU scores at 0.3, 0.5, and 0.7.


\section{Experiments and Discussion}
\begin{figure}[t]
\begin{center}
  \includegraphics[width=1.0\linewidth]{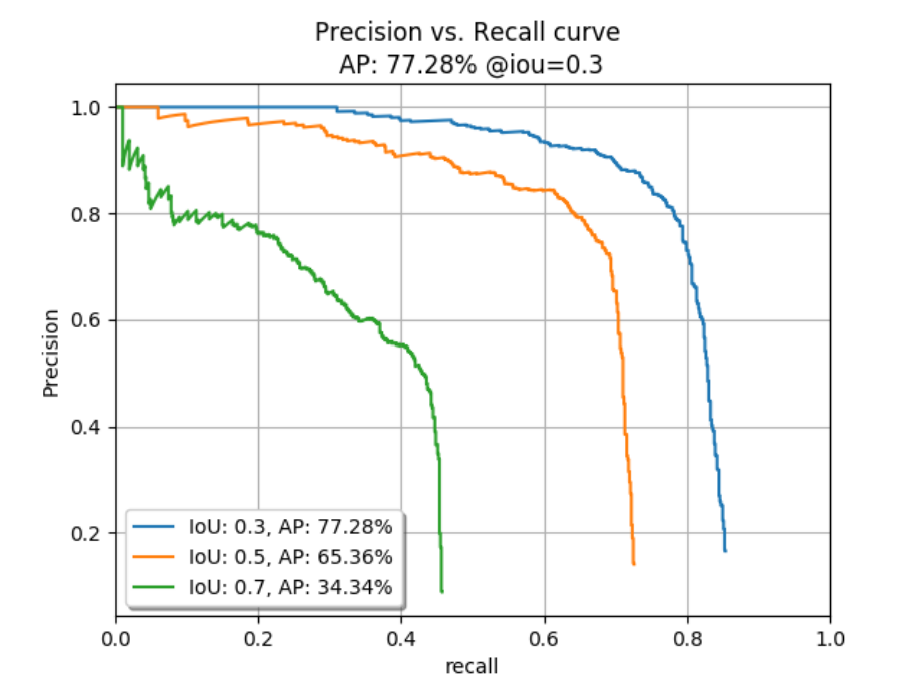}
\end{center}
  \caption{Performance of our detection algorithm}
\label{fig:complexyolo_map}
\end{figure}

\begin{figure*}[t]
\begin{center}
  \includegraphics[width=1.0\linewidth]{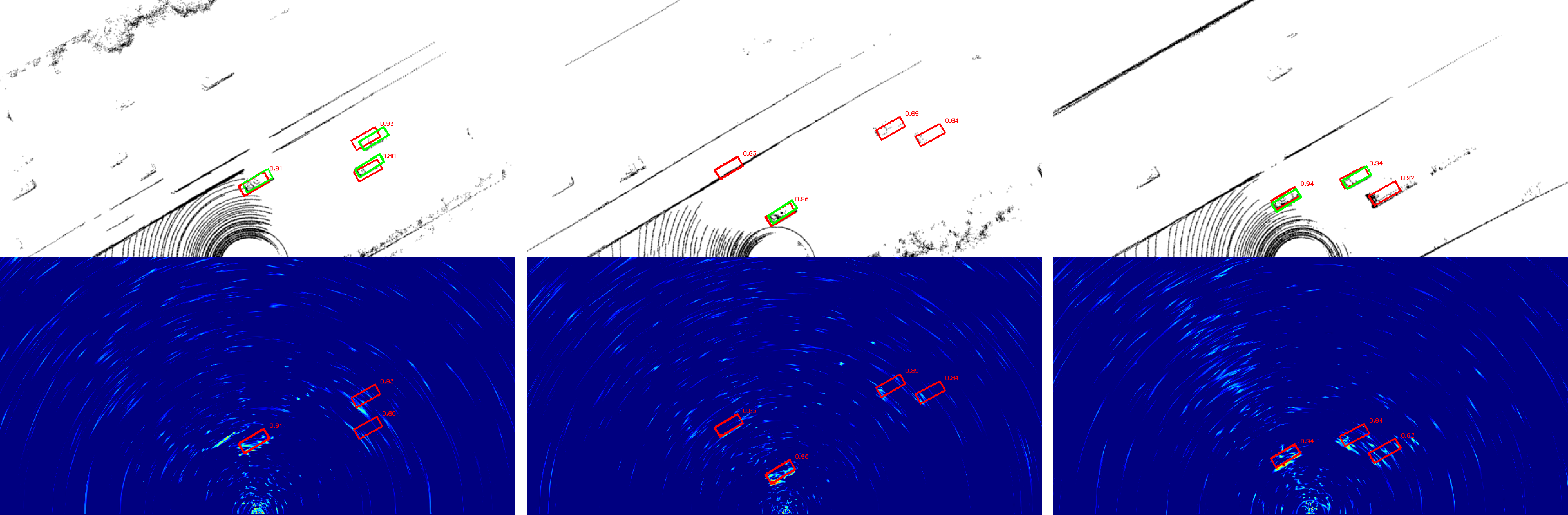}
\end{center}
  \caption{Three examples of predictions from our detection algorithm. In each of the three examples, the lower image is the radar image where the detector makes prediction upon, and the upper image is the corresponding LiDAR image for better visualization. The green boxes indicate the ground truth and the red boxes indicate the predictions. }
\label{fig:result_demo}
\end{figure*}

Fig. \ref{fig:complexyolo_map} shows the average precision (AP) result evaluated on highway scene. The best performance of AP as 77.28\% is achieved in our study (IoU at 0.3). Qualitative examples are shown in Fig. \ref{fig:result_demo}. Ablation studies were also conducted to investigate the effectiveness of different parts of the network.

\subsection{Radar Data Formats}
As mentioned in Fig. \ref{fig:radar_dsp}, four different radar data formats can be obtained from radar DSP, namely data-fft, data-music, image-fft, and image-music. We conducted experiments to find which data format is best suited for neural network to make prediction upon. For the data-fft and the data-music formats, as the data are in polar coordinate system where it is difficult for neural network to decode the shape of cars directly, we inserted a deterministic coordinate transformation layer right after the network backbone to transform feature map into Cartesian coordinate system before bbox prediction. The comparison results are shown in Table \ref{tab:data_repr_exp}, where we can see that the \textit{image-music} format achieves the best performance. This result indicates that neural network prefers more explicit (Cartesian coordinate system) and less noisy (with MUSIC processing) data format to perform object detection upon, which is in line with our expectation. 
\begin{table}[]
\caption{Comparison of radar data format}
\label{tab:data_repr_exp}
\begin{center}
\begin{tabular}{c|c}
\hline\hline
\textbf{Format}     &  \textbf{AP@IoU = 0.3/0.5/0.7}\\[1pt]
\hline
data-fft     & 64.60/49.90/23.75 \\[1pt]
data-music & 68.21/49.16/19.61\\[1pt]
image-fft & 73.66/57.56/29.80\\[1pt]
image-music & 76.66/65.63/31.39\\[1pt]
\hline\hline
\end{tabular}
\end{center}
\end{table}

\subsection{Data Augmentation and Auto-labeling Data}
Only a small portion of the dataset are human-labeled with good labeling quality, and the rest is auto-labeled by LiDAR object detection algorithms with relatively noisy ground truth information. Though the massive auto-labeling data can bring the benefit of big data, the noisy ground truth could also potentially hamper the performance. As a result, we conducted experiments to study effect of adding more noisy training data. Experiments were conducted by using human-labeled data alone, using human-labeled data along with data augmentation (image jittering and horizontal flipping), and using both human-labeled and auto-labeled data with data augmentation. The result is reported in Table \ref{tab:big_data_exp}, where we can see adding data augmentation and auto-labeled data can largely improve the algorithm performance, which reassures the power of big data for boosting deep learning algorithm performance.

\begin{table}[]
\caption{Data augmentation and auto-labeling data}
\label{tab:big_data_exp}
\begin{center}
\begin{tabular}{c|c}
\hline\hline
\textbf{Training data}     &  \textbf{AP@IoU = 0.3/0.5/0.7}\\[1pt]
\hline
human-labeled data only     & 70.66/52.75/26.45 \\[1pt]
+ data augmentation & 	73.76/53.88/29.19 \\[1pt]
+ auto-labeling data & 77.28/65.36/34.33\\[1pt]
\hline\hline
\end{tabular}
\end{center}
\end{table}

\subsection{Variance Prediction for Bounding Box}
To tackle the ambiguity in radar signal and potential noise in ground truth information, we introduced probabilistic model for bounding box localization regression. To study the effectiveness of the method, we did ablation study by removing the variance predictions so that the regression loss is reduced to normal $SL_1$ loss. As we can see in Table \ref{tab:variance_exp}, adding variance prediction is able to improve the performance of the object detector. The improvement is more pronounced under more stringent IoU threshold of 0.7. It is noteworthy that in autonomous driving, the uncertainty estimate itself is very useful for autonomous driving system's downstream components such as behavioral prediction and motion planning.

\begin{table}[]
\caption{Variance Prediction for Bounding Box Regression}
\label{tab:variance_exp}
\begin{center}
\begin{tabular}{c|c}
\hline\hline
\textbf{Regression}     &  \textbf{AP@IoU = 0.3/0.5/0.7}\\[1pt]
\hline 
w/o variance     & 76.81/63.57/31.53 \\[1pt]
w/ variance & 	77.28/65.36/34.33 \\[1pt]
\hline\hline
\end{tabular}
\end{center}
\end{table}

\subsection{Discussion}
Fig. \ref{fig:result_demo} displays qualitative examples of the network's predictions on the test dataset. In the middle subfigure, we can observe a false positive prediction caused by the highway barrier (the leftmost one). The other two ``false positives'' are actually caused by missing annotation in the ground truth. In the right subfigure, the network successfully detected one vehicle (a large truck) that is missing in the ground truth but failed to regress the correct size. This could be attributed to the lack of big trucks in the training dataset. 

One remedy for the missing detection of minority classes such as trucks could be improvement of LiDAR detection pipeline and the use of more anchors of different sizes during regression. Interestingly, though the network struggles to regress the correct size of the vehicles, it still correctly predicted the location of the back of the vehicle (\textit{i.e.}, nearest point of collision). It works in the favor of a safety-critical system such as autonomous driving.

Empirically, we found that the typical spatial extension for the network to perform well is within 30 meters in radial dimension and from 30$\degree$ to 150$\degree$ in azimuth dimension. Beyond this extension, due to the weak and noisy radar response, the network tends to fail being able to detect vehicles. A potential remedy could be utilizing hard-mining strategy during network training.

As directions for future work, we can integrate velocity measurement in our dataset as additional information to distinguish static and moving objects. This information could be leveraged by neural network to make better predictions. Another noteworthy point is that due to the strong location-specific noise patterns in radar signal, the noise pattern in transformed Cartesian image is not shift invariant. Na{\"i}ve data augmentation in general object detection on natural images will not work. Instead we could also introduce DSP-inspired data augmentation by introducing extra phase offset in frequency domain, which will lead to translational offset in spatial domain after FFT.

\section{Conclusions}
In this work, we proposed a radar tensor representation which is well-suited for CNN based object detection algorithm, and developed a scalable method to build up a radar dataset from scratch in such representation. A probabilistic object detection algorithm was developed to perform object detection directly on radar data. Average precision score under IoU at 0.3 is obtained to be 77.28 \%. Our results showed that by combining the power of DSP, big data, and convolutional neural networks, object detection based on radar data only is feasible and promising.

\section*{Acknowledgements}
We sincerely thank Huan Hu, Venkatesan N. Ekambaram, Eddi Sunarto, Raghav Inti, and Yuhao Shui for insightful discussions and hardware support.

{\small
\bibliographystyle{ieee_fullname}
\bibliography{egbib}

\begin{thebibliography}{10}\itemsep=-1pt

\bibitem{monoloco}
Lorenzo Bertoni, Sven Kreiss, and Alexandre Alahi.
\newblock Monoloco: Monocular 3d pedestrian localization and uncertainty
  estimation.
\newblock In {\em The IEEE International Conference on Computer Vision (ICCV)},
  October 2019.

\bibitem{bijelic2018benchmark}
Mario Bijelic, Tobias Gruber, and Werner Ritter.
\newblock A benchmark for lidar sensors in fog: Is detection breaking down?
\newblock In {\em 2018 IEEE Intelligent Vehicles Symposium (IV)}, pages
  760--767. IEEE, 2018.

\bibitem{brodeski2019deep}
Daniel Brodeski, Igal Bilik, and Raja Giryes.
\newblock Deep radar detector, 2019.

\bibitem{nuscenes}
Holger Caesar, Varun Bankiti, Alex~H. Lang, Sourabh Vora, Venice~Erin Liong,
  Qiang Xu, Anush Krishnan, Yu Pan, Giancarlo Baldan, and Oscar Beijbom.
\newblock nuscenes: A multimodal dataset for autonomous driving.
\newblock {\em arXiv preprint arXiv:1903.11027}, 2019.

\bibitem{cascade-rcnn}
Zhaowei Cai and Nuno Vasconcelos.
\newblock Cascade r-cnn: Delving into high quality object detection, 2017.

\bibitem{gaussianYOLO}
Jiwoong Choi, Dayoung Chun, Hyun Kim, and Hyuk-Jae Lee.
\newblock Gaussian yolov3: An accurate and fast object detector using
  localization uncertainty for autonomous driving, 2019.

\bibitem{R-FCN}
Jifeng Dai, Yi Li, Kaiming He, and Jian Sun.
\newblock R-fcn: Object detection via region-based fully convolutional
  networks, 2016.

\bibitem{dcnn}
Jifeng Dai, Haozhi Qi, Yuwen Xiong, Yi Li, Guodong Zhang, Han Hu, and Yichen
  Wei.
\newblock Deformable convolutional networks, 2017.

\bibitem{centernet-2}
Kaiwen Duan, Song Bai, Lingxi Xie, Honggang Qi, Qingming Huang, and Qi Tian.
\newblock Centernet: Keypoint triplets for object detection, 2019.

\bibitem{cfar}
H.~M. FINN.
\newblock Adaptive detection mode with threshold control as a function of
  spatially sampled clutter level estimates.
\newblock {\em RCA Rev.}, 29:414--463, 1968.

\bibitem{furukawa2018deep}
Hidetoshi Furukawa.
\newblock Deep learning for end-to-end automatic target recognition from
  synthetic aperture radar imagery, 2018.

\bibitem{fast-rcnn}
Ross Girshick.
\newblock Fast r-cnn, 2015.

\bibitem{rcnn}
Ross Girshick, Jeff Donahue, Trevor Darrell, and Jitendra Malik.
\newblock Rich feature hierarchies for accurate object detection and semantic
  segmentation, 2013.

\bibitem{resnet}
Kaiming He, Xiangyu Zhang, Shaoqing Ren, and Jian Sun.
\newblock Deep residual learning for image recognition, 2015.

\bibitem{he2019bounding}
Yihui He, Chenchen Zhu, Jianren Wang, Marios Savvides, and Xiangyu Zhang.
\newblock Bounding box regression with uncertainty for accurate object
  detection.
\newblock In {\em Proceedings of the IEEE Conference on Computer Vision and
  Pattern Recognition}, pages 2888--2897, 2019.

\bibitem{spatial-transformer}
Max Jaderberg, Karen Simonyan, Andrew Zisserman, and Koray Kavukcuoglu.
\newblock Spatial transformer networks, 2015.

\bibitem{jiang2017r2cnn}
Yingying Jiang, Xiangyu Zhu, Xiaobing Wang, Shuli Yang, Wei Li, Hua Wang, Pei
  Fu, and Zhenbo Luo.
\newblock R2cnn: rotational region cnn for orientation robust scene text
  detection.
\newblock {\em arXiv preprint arXiv:1706.09579}, 2017.

\bibitem{uncertainties}
Alex Kendall and Yarin Gal.
\newblock What uncertainties do we need in bayesian deep learning for computer
  vision?, 2017.

\bibitem{DFP}
Tao Kong, Fuchun Sun, Wenbing Huang, and Huaping Liu.
\newblock Deep feature pyramid reconfiguration for object detection, 2018.

\bibitem{foveabox}
Tao Kong, Fuchun Sun, Huaping Liu, Yuning Jiang, and Jianbo Shi.
\newblock Foveabox: Beyond anchor-based object detector, 2019.

\bibitem{fast-cfar}
M. {Kronauge} and H. {Rohling}.
\newblock Fast two-dimensional cfar procedure.
\newblock {\em IEEE Transactions on Aerospace and Electronic Systems},
  49(3):1817--1823, July 2013.

\bibitem{kutila2018automotive}
Matti Kutila, Pasi Pyyk{\"o}nen, Hanno Holzh{\"u}ter, Mich{\`e}le Colomb, and
  Pierre Duthon.
\newblock Automotive lidar performance verification in fog and rain.
\newblock In {\em 2018 21st International Conference on Intelligent
  Transportation Systems (ITSC)}, pages 1695--1701. IEEE, 2018.

\bibitem{acc_radar_sim}
C. {Kärnfelt}, A. {Péden}, A. {Bazzi}, G. {El Haj Shhadé}, M. {Abbas}, and
  T. {Chonavel}.
\newblock 77 ghz acc radar simulation platform.
\newblock In {\em 2009 9th International Conference on Intelligent Transport
  Systems Telecommunications, (ITST)}, pages 209--214, Oct 2009.

\bibitem{cornernet}
Hei Law and Jia Deng.
\newblock Cornernet: Detecting objects as paired keypoints, 2018.

\bibitem{fpn}
Tsung-Yi Lin, Piotr Dollár, Ross Girshick, Kaiming He, Bharath Hariharan, and
  Serge Belongie.
\newblock Feature pyramid networks for object detection, 2016.

\bibitem{retinanet}
Tsung-Yi Lin, Priya Goyal, Ross Girshick, Kaiming He, and Piotr Dollár.
\newblock Focal loss for dense object detection, 2017.

\bibitem{ssd}
Wei Liu, Dragomir Anguelov, Dumitru Erhan, Christian Szegedy, Scott Reed,
  Cheng-Yang Fu, and Alexander~C. Berg.
\newblock Ssd: Single shot multibox detector.
\newblock {\em Lecture Notes in Computer Science}, page 21–37, 2016.

\bibitem{liu2016ship}
Zikun Liu, Hongzhen Wang, Lubin Weng, and Yiping Yang.
\newblock Ship rotated bounding box space for ship extraction from
  high-resolution optical satellite images with complex backgrounds.
\newblock {\em IEEE Geoscience and Remote Sensing Letters}, 13(8):1074--1078,
  2016.

\bibitem{cosinelr}
Ilya Loshchilov and Frank Hutter.
\newblock Sgdr: Stochastic gradient descent with warm restarts, 2016.

\bibitem{ma2018arbitrary}
Jianqi Ma, Weiyuan Shao, Hao Ye, Li Wang, Hong Wang, Yingbin Zheng, and
  Xiangyang Xue.
\newblock Arbitrary-oriented scene text detection via rotation proposals.
\newblock {\em IEEE Transactions on Multimedia}, 20(11):3111--3122, 2018.

\bibitem{Ma_2018}
Jianqi Ma, Weiyuan Shao, Hao Ye, Li Wang, Hong Wang, Yingbin Zheng, and
  Xiangyang Xue.
\newblock Arbitrary-oriented scene text detection via rotation proposals.
\newblock {\em IEEE Transactions on Multimedia}, 20(11):3111–3122, Nov 2018.

\bibitem{massa2016crafting}
Francisco Massa, Renaud Marlet, and Mathieu Aubry.
\newblock Crafting a multi-task cnn for viewpoint estimation, 2016.

\bibitem{lasernet}
Gregory~P. Meyer, Ankit Laddha, Eric Kee, Carlos Vallespi-Gonzalez, and Carl~K.
  Wellington.
\newblock Lasernet: An efficient probabilistic 3d object detector for
  autonomous driving, 2019.

\bibitem{astyx_dataset}
M. {Meyer} and G. {Kuschk}.
\newblock Automotive radar dataset for deep learning based 3d object detection.
\newblock In {\em 2019 16th European Radar Conference (EuRAD)}, pages 129--132,
  Oct 2019.

\bibitem{mousavian20163d}
Arsalan Mousavian, Dragomir Anguelov, John Flynn, and Jana Kosecka.
\newblock 3d bounding box estimation using deep learning and geometry, 2016.

\bibitem{Neven_2019_CVPR}
Davy Neven, Bert~De Brabandere, Marc Proesmans, and Luc~Van Gool.
\newblock Instance segmentation by jointly optimizing spatial embeddings and
  clustering bandwidth.
\newblock In {\em The IEEE Conference on Computer Vision and Pattern
  Recognition (CVPR)}, June 2019.

\bibitem{Patel2019DeepLO}
Kanil Patel, Kilian Rambach, Tristan Visentin, Daniel Rusev, Michael Pfeiffer,
  and Bin Yang.
\newblock Deep learning-based object classification on automotive radar
  spectra.
\newblock {\em 2019 IEEE Radar Conference (RadarConf)}, pages 1--6, 2019.

\bibitem{yolov3}
Joseph Redmon and Ali Farhadi.
\newblock Yolov3: An incremental improvement.
\newblock {\em CoRR}, abs/1804.02767, 2018.

\bibitem{faster-rcnn}
Shaoqing Ren, Kaiming He, Ross Girshick, and Jian Sun.
\newblock Faster r-cnn: Towards real-time object detection with region proposal
  networks, 2015.

\bibitem{esprit}
R. {Roy} and T. {Kailath}.
\newblock Esprit-estimation of signal parameters via rotational invariance
  techniques.
\newblock {\em IEEE Transactions on Acoustics, Speech, and Signal Processing},
  37(7):984--995, July 1989.

\bibitem{music}
R. {Schmidt}.
\newblock Multiple emitter location and signal parameter estimation.
\newblock {\em IEEE Transactions on Antennas and Propagation}, 34(3):276--280,
  March 1986.

\bibitem{pointrcnn}
Shaoshuai Shi, Xiaogang Wang, and Hongsheng Li.
\newblock Pointrcnn: 3d object proposal generation and detection from point
  cloud, 2018.

\bibitem{shnidman2003expanded}
DA Shnidman.
\newblock Expanded swerling target models.
\newblock {\em IEEE Transactions on Aerospace and Electronic Systems},
  39(3):1059--1069, 2003.

\bibitem{complexyolo}
Martin Simony, Stefan Milzy, Karl Amendey, and Horst-Michael Gross.
\newblock Complex-yolo: An euler-region-proposal for real-time 3d object
  detection on point clouds.
\newblock In {\em The European Conference on Computer Vision (ECCV) Workshops},
  September 2018.

\bibitem{wang2019pseudo}
Yan Wang, Wei-Lun Chao, Divyansh Garg, Bharath Hariharan, Mark Campbell, and
  Kilian~Q Weinberger.
\newblock Pseudo-lidar from visual depth estimation: Bridging the gap in 3d
  object detection for autonomous driving.
\newblock In {\em Proceedings of the IEEE Conference on Computer Vision and
  Pattern Recognition}, pages 8445--8453, 2019.

\bibitem{xia2017dota}
Gui-Song Xia, Xiang Bai, Jian Ding, Zhen Zhu, Serge Belongie, Jiebo Luo, Mihai
  Datcu, Marcello Pelillo, and Liangpei Zhang.
\newblock Dota: A large-scale dataset for object detection in aerial images,
  2017.

\bibitem{reppoints}
Ze Yang, Shaohui Liu, Han Hu, Liwei Wang, and Stephen Lin.
\newblock Reppoints: Point set representation for object detection, 2019.

\bibitem{STDN}
Peng Zhou, Bingbing Ni, Cong Geng, Jianguo Hu, and Yi Xu.
\newblock Scale-transferrable object detection.
\newblock In {\em proceedings of the IEEE conference on computer vision and
  pattern recognition}, pages 528--537, 2018.

\bibitem{centernet}
Xingyi Zhou, Dequan Wang, and Philipp Krähenbühl.
\newblock Objects as points, 2019.

\bibitem{extremenet}
Xingyi Zhou, Jiacheng Zhuo, and Philipp Kr{\"{a}}henb{\"{u}}hl.
\newblock Bottom-up object detection by grouping extreme and center points.
\newblock {\em CoRR}, abs/1901.08043, 2019.

\bibitem{dcnn-v2}
Xizhou Zhu, Han Hu, Stephen Lin, and Jifeng Dai.
\newblock Deformable convnets v2: More deformable, better results, 2018.

\end{thebibliography}
}

\end{document}